\documentclass[12pt, letterpaper]{article}
\usepackage[utf8x]{inputenc} 
\usepackage[spanish, mexico]{babel} 
\usepackage[T1]{fontenc}
\usepackage{amssymb}
\usepackage{mathtools}
\usepackage[usenames]{color}
\usepackage{float}
\usepackage{listings}
\usepackage{graphicx}
\usepackage{graphicx}  
\usepackage[left=2.5cm,top=2.0cm,right=2cm,bottom=2.0cm]{geometry}  
\begin{document}

\definecolor{col}{RGB}{230,230,230}

\title{\textbf{Diseño de un espacio semántico sobre la base de la Wikipedia.}\\
\Large{\textit{ Una propuesta de análisis de la semántica latente para el idioma español.}}}

\author{Dalina Aidee Villa$^{a}$, Igor Barahona$^{b}$, Luis Javier Álvarez $^{b}$\\
\small{$^{a}$ Facultad de Ciencias, Universidad Nacional Autónoma de México,}\\
\small{$^{b}$ Instituto de Matemáticas, Unidad Cuernavaca, Universidad Nacional Autónoma de México}}

\date{enero, 2019}

\maketitle

\vspace*{-1cm}
\begin{center}\rule{0.9\textwidth}{0.1mm} \end{center}
\begin{abstract}

 El análisis de Semántica Latente (LSA, por sus siglas en inglés) fue concebido inicialmente desde el área de la psicología cognitiva.  Desde su aparición en la década de los 90s, el LSA ha sido utilizado para modelar procesos cognitivos, corregir textos académicos, comparar obras literarias y analizar discursos, entre otras. 
Sobre la base de métodos estadísticos multivariados de reducción de dimensionalidad, aplicados a la Wikipedia en texto plano, este trabajo propone la construcción de un espacio semántico (diccionario) para el idioma Español. Nuestros resultados incluyen una matriz documentos palabras (con dimensiones de $1.3 X10^{6} $renglones y $5.9 X10^{6} $ columnas). Los valores singulares de tal matriz, son utilizados para comparar semánticamente dos o más textos.\\

\textbf{Palabras clave:} “análisis semántico”, “espacio semántico”, “reducción de dimensiones”,  “valores singulares”, “Estadística multivariada”
\begin{center}\rule{0.9\textwidth}{0.1mm} \end{center}
\vspace*{0.5cm}
\end{abstract}

 \section*{Introducción}
  Desde las primeras civilizaciones que existieron, la raza humana siempre ha mostrado una necesidad imperiosa por comunicarse.  El lenguaje es el instrumento de comunicación más efectivo que haya conocido la humanidad. Una interpretación correcta del lenguaje es un requisito indispensable dentro del proceso de comunicación efectivo. Los campos semánticos, los cuales están basados en métodos estadísticos multivariados, han resultado una poderosa herramienta para lograr una interpretación correcta de las palabras de un idioma.\\
   
\section*{Contexto}

El análisis de la semántica latente (LSA, por sus siglas en inglés) fue concebido inicialmente desde el área de la psicología cognitiva, como una técnica que fuera capaz de adquirir y recuperar el conocimiento humano (Deerwester et al 1990). Desde sus primeras apariciones en la década de los 90s, el LSA ha sido utilizado en diferentes áreas del conocimiento, entre las que se pueden destacar: la modelación de procesos cognitivos, corrección de textos académicos, análisis del discurso, obtención de medidas de cohesión textual y comparación otras literarias, entre otras (Landauer, 1999; 
Kintsch,1998; Kintsch, 2001; Kintsch y Bowles, 2002). 
En el campo del análisis estadístico de textos, el LSA ha demostrado su precisión para describir correctamente las relaciones que mantienen unas palabras con otras en diferentes contextos. Las medidas cuantitativas de las relaciones de palabras similares, utilizadas en diferentes contextos, nos permite crear conceptos o significados semánticos. Además, tales medidas sientan las bases para la creación de tópicos y redes temáticas. Es decir, a través de la aplicación de métodos estadísticos multivariados, el LSA pone de manifiesto una propiedad que se cumple en los lenguajes naturales: aquellas palabras valores semánticos similares, aparecerán en contextos parecidos.\\

La idea central que soporta el análisis semántico latente es relativamente sencilla: cualquier texto, sin importar el tamaño puede ser representado por un sistema de ecuaciones lineales.  En este sistema, el significado de alguna palabra en particular estará representado por la suma de los documentos (contextos) donde es utilizada.  De lo anterior, se desprende que dos o más palabras tendrán un significado semántico similar, el cual estará en relación directa con sus respectivas sumas.  La técnica de descomposición de valores singulares, es utilizada como un método de reducción de dimensionalidad. Esto nos permite crear un espacio semántico de menores dimensiones, el cual es una representación de las palabras en función de sus eigenvectores.  Este procedimiento será explicado con mayor detalle en las siguientes páginas.

\section*{Objetivo}
El objetivo de este trabajo es proponer un campo semántico para el idioma español, tomando como punto de partida la Wikipedia en este idioma. Tal campo semántico deberá incluir las relaciones latentes que existen entre todas las palabras que componen el idioma español, o al menos aquellas representadas en la Wikipedia. Las razones para seleccionar esta base de datos y no otra, es que la Wikipedia al ser un proyecto colaborativo, hace posible que todas las personas con acceso a internet e hispanoparlantes sean potenciales colaboradores. Lo anterior nos permite cumplir en forma parcial con el supuesto, de que un campo semántico, deberá ser construido sobre de textos en lenguaje natural. En cuanto su contribución, se espera que este trabajo haga una aportación original a las ciencias sociales, en el sentido de proponer herramientas con rigor matemático, las cuales facilitan los procesos de toma de decisiones. \\

\section*{Materiales y métodos}

En el contexto del análisis semántico latente, resulta de especial importancia que los textos utilizados se compongan de lenguaje natural.  Las bases de datos bajo análisis se deben componer del idioma español escrito, el cual es utilizado para propósitos generales de comunicación.  Otra característica por observar en los textos analizados es que estos deben ser tan extensos, como sea posible. Tal como se mencionó anteriormente, un inconveniente en el Análisis Semántico Latente (LSA) es que la precisión de los resultados a obtener se encuentra en función del tamaño del texto investigado. Tal como se menciona en Günther, Dudschig y Kaup (2015), la lógica detrás del LSA es que, a mayor tamaño de las bases de datos utilizadas, mayor la precisión de los resultados. La Wikipedia ofrece algunos beneficios en esa dirección, tal como lo ilustramos a continuación.\\\\

La Wikipedia es una enciclopedia libre, editada de forma colaborativa en más de 300 idiomas.  Es administrada por Fundación Wikipedia, una organización sin fines de lucro y cuya financiación se basa en las donaciones de los usuarios (Wikipedia, 2018). A la fecha de elaboración del presente documento, se estima que la Wikipedia en español tenía aproximadamente 48 millones de artículos, los cuales son redactados por voluntarios alrededor del mundo. Desde su creación en el año 2001, el proyecto ha recibido cerca de 2,000 millones de ediciones. Un factor crítico en para el éxito de la Wikipedia es que cualquier persona con una computadora y acceso a internet puede sumarse al proyecto, a través de realizar ediciones, cambios, aportaciones, etc.  Por lo anterior, consideramos que la Wikipedia cumplía con las características necesarias para realizar LSA.  En primer lugar, al ser editada por muchos usuarios con diferentes perfiles sociodemográficos, se cumple parcialmente el requisito del lenguaje natural. Aun cuando tiene sesgos importantes sobre la forma y uso del lenguaje, la Wikipedia es uno de los proyectos digitales más importantes que existen para el idioma español. Esto último representa una ventaja al momento de construir el campo semántico, porque se evitan procesos de digitalización que quedan fuera del alcance de este trabajo. Para sopesar las limitantes antes mencionadas, un trabajo futuro tomará como punto de partida bibliotecas más extensas que la Wikipedia, pero que no están totalmente digitalizadas. \\\\

La metodología implementada en este trabajo se compone de siete pasos, comenzando con la descarga de la Wikipedia en texto plano y finalizando con una ilustración de la comparación semántica de textos (ver Figura 1).  En los siguientes renglones explica cada uno de ellos. \\

\begin{figure}[H]
\centering
\includegraphics[width=.60\textwidth]{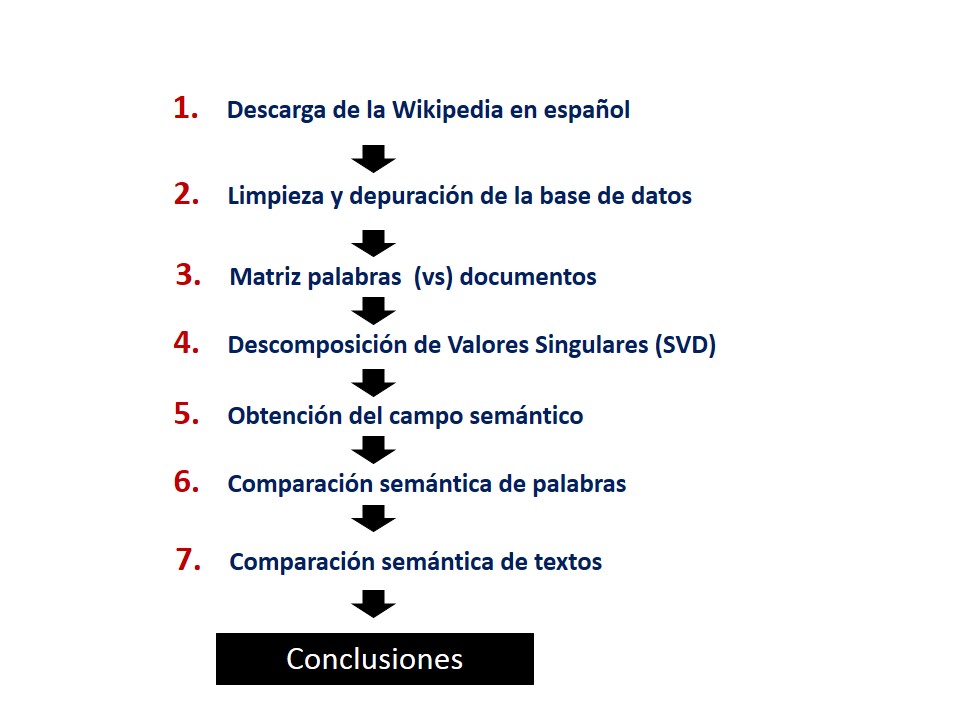}
\caption{Diagrama de Flujo}
\label{Diagrama de Flujo}
\end{figure}

\subsubsection*{1. Descargar la Wikipedia}

El punto de partida es descargar del sitio $ https://www.wikipedia.org/ $ la Wikipedia en texto plano y sin imágenes.  El archivo descargado tiene un formato XML y un peso de 2.5 GB.  De acuerdo con Wikipedia (2018) estos documentos son actualizados mensualmente y para fines de este trabajo hemos tomado la versión de octubre 2018.  La libredia “WikiExtractor”, que es un script en Python que extrae y limpia el texto desde un archivo de Wikipedia tipo “dump”, fue utilizado para este propósito (WikiEstracto, 2018). Mediante este algoritmo, un total de 7,793,289 documentos que corresponden al mismo número de páginas fueron almacenados en el disco local. \\\\

 \subsubsection{2. Limpieza y depuración de bases de datos}

A diferencia de otras bases de datos digitales, los contenidos tipo texto de la Wikipedia se distribuyen bajo la licencia “Creative Commons Attribution-ShareAlike 3.0” $(ver:  https://creativecommons.org )$. Lo anterior permite utilizarla para copias de seguridad, consultas, uso personal o proyectos académicos. Una copia en formato XML y en idioma español fue descargada de la página de Wikipedia (ver: $https://dumps.wikimedia.org/)$ y posteriormente realizamos tareas de limpieza y depuración. Específicamente, se aplicó el procedimiento sugerido por Attardi, Cozza y Sartiano (2015), el cual consiste en remover hipervínculos, caracteres especiales, puntuación y signos especiales. Con lo anterior obtuvimos un conjunto de datos con 1,336,063 renglones y tres columnas. Estas últimas hacen referencia al identificador del documento, la palabra clave y su respectiva definición. \\\\

\subsubsection*{3. Construir una matriz documentos-palabras}

Un elemento de gran importancia en el análisis semántico es la matriz $\textbf{\textit{M}}$ o Matriz Documentos-Palabras (DTM). Se dice que $\textbf{\textit{M}}$ tiene dimensiones $\textbf{\textit{n}}$ x $\textbf{\textit{p}}$, donde $\textbf{\textit{i}}$=1,..,$\textbf{\textit{n}}$ representa el número total de páginas de las que se compone la Wikipedia y $\textbf{\textit{j}}$=1,…,$\textbf{\textit{p}}$ hace referencia al total de palabras únicas.  Además \textbf{\textit{$x_{i,j}$}} denota el número de veces que la palabra $\textbf{\textit{i}}$ aparece en el documento $\textbf{\textit{j}}$, por lo tanto, $\textbf{\textit{M}}$ también puede ser considerada una matriz de frecuencias. Las librerías “tm”, “Matrix” y “lsa”, creadas bajo el ambiente de R fueron utilizadas durante la construcción de la DTM.  Adicionalmente, se concatenaron las variables ID, ID-familia y letra del abecedario como suplementarias al estudio, tal como se ilustra en la Figura 2. \\\\
 
 \begin{figure}[H]
\centering
\includegraphics[width=.60\textwidth]{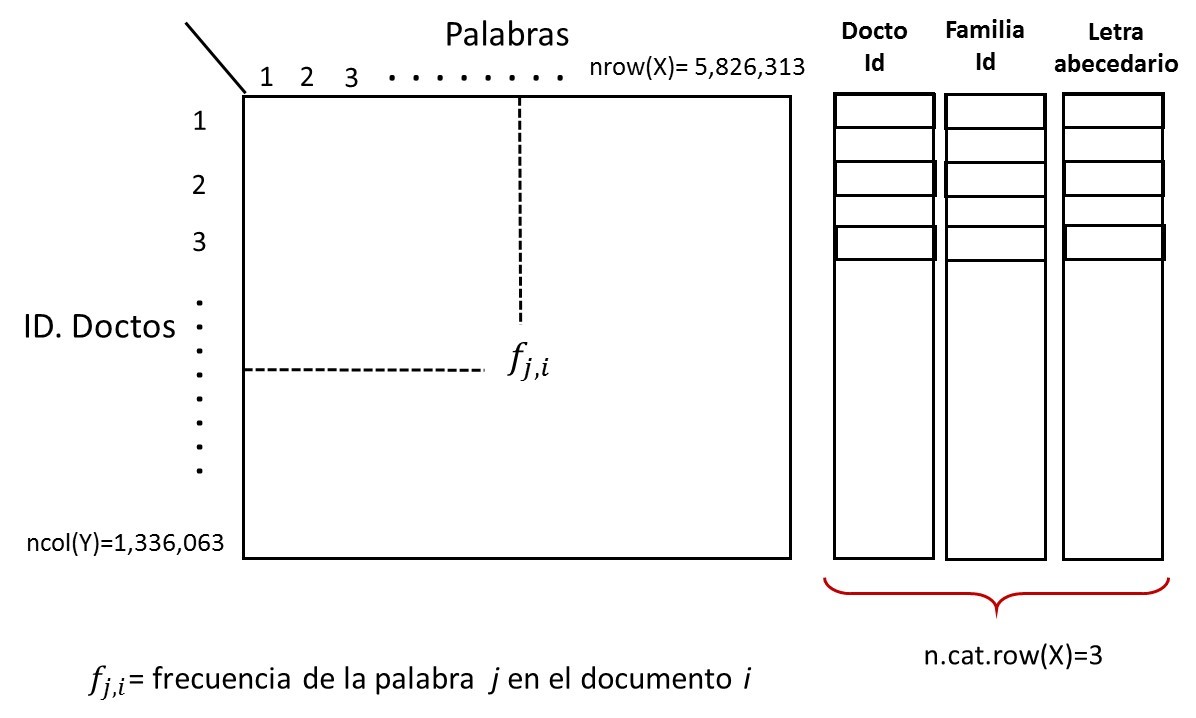}
\caption{Matriz documentos palabras}
\label{Matriz Documentos palabras}
\end{figure}

 Dentro de la carga de información se obtuvo lo siguiente que fue la representación de la información. 

\lstset{language=R, breaklines=true, basicstyle=\footnotesize}
\lstset{numbers=right, numberstyle=\tiny, stepnumber=1, numbersep=-2pt}
\begin{lstlisting}[frame=single]
load("mi.lsaV2.rda")
\end{lstlisting}

El procesamiento de los datos se dio en el software R.\\

\subsubsection*{4. Descomposición Singular}

Sea \textbf{\textit{M}} una matriz de orden $\textbf{\textit{n}}$ x $\textbf{\textit{p}}$ con \textit{n} $ \geq$ \textit{p}. Se llama descomposición de valores singulares de \textbf{\textit{M}} a: \\

\begin{equation}
\textbf{\textit{M}}=UDV^{T}
\end{equation}\\

Donde \textbf{\textit{U}} es la matriz de \textit{n} x \textit{p} cuyas columnas son vectores ortogonales , \textbf{\textit{D}} es una matriz diagonal \textit{p} x \textit{p}. Además \textbf{\textit{V}} es una matriz de \textit{p} x \textit{p} ortogonal.\\

Se verifica que:\\

\begin{itemize}

\item[1] El rango de  \textbf{\textit{M}} es el número de \textit{r} de valores singulares positivos.\\

\item[2]  \textbf{\textit{U}} contiene a los vectores propios de  \textbf{\textit{M}}  \textbf{\textit{$M^{T}$}}, siendo igual  \textbf{\textit{M}}  \textbf{\textit{$M^{T}$}}

\item[3] \textbf{\textit{V}} contiene los vectores propios de  \textbf{\textit{M}}  \textbf{\textit{$M^{T}$}}, siendo   \textbf{\textit{$V^{T}$}} \textbf{\textit{V}} $=$  \textbf{\textit{V}}\textbf{\textit{$V^{T}$}} $=I$

\item[4] Si \textit{n} $=$ \textit{p} y \textbf{\textit{M}} es simétrica, entonces \textbf{\textit{U}} $=$ \textbf{\textit{V}} y  \textbf{\textit{M}}$= UDU^{T}$ es la descomposición  espectral de \textbf{\textit{M}}. Los valores singulares son los valores propios de \textbf{\textit{M}}.

\end{itemize}

La matriz obtenida en el punto anterior es una matriz dispersa. De acuerdo con Cuadras (2014), se considera una matriz dispersa aquella de gran tamaño, y en donde la mayor parte de sus elementos son igual a cero. El número de elementos igual a cero requeridos para considerar a una matriz como dispersa, depende de su propia estructura, así como de las operaciones que se realizarán con la misma. El algoritmo propuesto por Lanczos (1950) y revisado por Berry (1992) fueron utilizados para hacer la SVD en la matriz dispersa $\textbf{\textit{M}}$. Básicamente, este procedimiento genera una secuencia de $T_{,j}$ matrices tri-diagonales, las cuales tienen la propiedad de que los valores propios extremos de cada matriz $T_{,j}$, con dimensiones $\textit{j}$ x $\textit{j}$  son las mejores estimaciones de los valores propios de la matriz original. Por ejemplo, la matriz original  \textbf{\textit{M}} puede expresarse de acuerdo a la siguiente formulación:\\\\. 

\begin{equation}
\textbf{\textit{M}}= \begin{pmatrix}0&A\\A^{T}&0\end{pmatrix} 
\end{equation}\\

En donde la matriz $\textit{A}$ es simétrica con dimensiones ( $\textit{n}$ x $\textit{p}$) x ( $\textit{n}$ x $\textit{p}$), y además $\textit{n}$ > $\textit{p}$. Entonces, si el rango($\textit{A}$) = $\textit{n}$, se puede observar que los valores propios de  $\textbf{\textit{M}}$ están dados por los $\textit{n}$ pares,  $\pm\sigma_{i}$, donde $\sigma_{i}$ es un valor singular de $\textit{A}$, con ($\textit{p}$-$\textit{n}$) valores singulares igual a cero.  Además, la multiplicidad de los valores singulares igual a cero en la matriz original  $\textbf{\textit{M}}$ es igual a $\textit{n}$+$\textit{p}$-2$\textit{r}$, donde $\textit{r}$=rango($\textit{A}$).  Posteriormente, Rohde et. al. (2018) implementaron tal procedimiento en la librería de R “sparsesvd”, ésta última utilizada para obtener los resultados que explicamos en la siguiente sección. \\\\

Por último, previo a la descomposición en valores singulares de la matriz $\textbf{\textit{M}}$, se requiere restar importancia a las palabras con una frecuencia muy alta, y al mismo tiempo aumentar este valor para aquellas con frecuencias muy bajas. De acuerdo con Nakov, Popova y Mateev, (2001) aquellas palabras con las frecuencias más altas en la matriz \textbf{\textit{M}}, resultan inservibles para discriminar con precisión el valor semántico proporcionado en cada documento \textit{i}, mientras usar aquellas con valores en sus frecuencias cercanos a la media resultan en mayores niveles de precisión. \\

En Barahona (2016) se menciona que la técnica de SVD es una generalización del análisis de componentes principales y una variante del bien conocido análisis factorial.  Este consiste en descomponer la matriz \textbf{\textit{M}} en tres elementos, tal como se muestra a continuación:\\\

\begin{equation}
 \textbf{\textit{$M_{d}$}}= UDV^{T}
\end{equation}\\

De acuerdo a la formulación (3)  \textbf{\textit{$M_{d}$}} es un ajuste obtenido a partir de la matriz original  \textbf{\textit{M}}. Además \textbf{\textit{U}} es una matriz ortogonal con dimensiones \textbf{\textit{j}} x \textbf{\textit{r}}; \textbf{\textit{$V^{T}$}} es también una matriz ortogonal con dimensiones \textbf{\textit{i}} x  \textbf{\textit{r}}.  Por último,  \textbf{\textit{ D}} representa la matriz diagonal. Nótese que  \textbf{\textit{ r }} está dado por el orden la matriz original  \textbf{\textit{M}}. Ahora realizamos una reducción de las dimensiones en \textbf{\textit{ U}} y \textbf{\textit{$ V^{T}$}}. Es decir, si ambas matrices son del orden \textbf{\textit{r}}, entonces son reducidas a \textbf{\textit{k}} dimensiones,  donde \textbf{\textit{k < r}}. De acuerdo a Landauer y Dumais (1997) \textbf{\textit{k}} es arbitrario y valor dependerá del conjunto de datos inicial. Estudios empíricos sugieren valores para \textbf{\textit{k}} de alrededor de 300.\\

El algoritmo explicado anteriormente nos permite representar documentos y palabras, a través de promedios obtenidos por los  \textbf{\textit{k}} vectores retenidos de la matriz  \textbf{\textit{$M_{d}$}}. De esta manera, dos palabras tendrán un significado semántico parecido, si las medias de sus vectores son similares. Para realizar comparaciones entre documentos y palabras, calculamos el coseno del ángulo entre los vectores de interés. Por ejemplo, un coseno igual a 0 entre dos palabras nos indicará ortogonalidad, y por lo tanto, una diferencia semántica importante entre ambas. Por el contrario, entre más cercano a 1.0 sea el coseno entre vectores, nos indicará una similitud semántica mayor. Este tipo de investigaciones son bien conocidas para los idiomas Inglés, Francés, Alemán y Serbio y en el contexto latinoamericano este trabajo resulta una aportación novedosa. \\

\subsubsection*{5.Obtención del campo semántico}

Debido a que los documentos analizados son de diferentes extensiones, resulta interesante hacer una comparación de dos campos semánticos. El primero obtenido a partir una matriz de frecuencias  \textbf{\textit{M}}, con dimensiones  $\textbf{\textit{n}}$ x  $\textbf{\textit{p}}$,  donde $\textit{i}$=1,…,$\textbf{\textit{n}}$ hace referencia al número de documentos y $\textit{j}$=1,…,$\textbf{\textit{p}}$ al total de palabras consideradas en el análisis. Por lo tanto \textbf{\textit{$x_{i,j}$}} representa la frecuencia con la que aparece la palabra $\textbf{\textit{p}}$ en el documento $\textbf{\textit{n}}$.  El segundo obtenido a partir de una matriz binaria con las mismas dimensiones, en la cual cada elemento \textbf{\textit{$x_{i,j}$}} es igual cero en los casos en que la palabra $\textbf{\textit{p}}$ aparece cero veces en el documento $\textbf{\textit{n}}$, en caso contrario es igual a uno.  En ambos casos de retuvieron 300 dimensiones, tal como se sugiere en la literatura. \\

Como resultado de tal comparación, se encontraron diferencias importantes entre ambos campos semánticos.  Las matrices diagonales de ambos campos tienden a ser muy diferentes en los extremos y similares en la parte central, presentando el primero valores superiores en contraste con el segundo. De lo anterior se deduce que el campo semántico construido a partir de la matriz de frecuencias es más sensible a variaciones semánticas, en contraste con el elaborado a partir de la matriz binaria.\\

Durante el proceso antes descrito, utilizamos la librería “sparsesvd” (Rohde et. al 2018), la cual está diseñada sobre la base del algoritmo propuesto por Lanczos (1950) para realizar el SVD en matrices dispersas. Así también, con la finalidad de ponderar la matriz documentos palabras, seguimos el algoritmo sugerido por Günther, Dudschig and Kaup (2014). La ejecución numérica de esta ponderación se realizó a través de la librería “lsa” (Wild, 2015).\\

\subsubsection*{6.Comparación semántica de palabras}
Tomando en cuenta que cada palabra es identificada con un vector único dentro del campo semántico, la similitud entre dos o más palabras estará dada por el coseno de sus respectivos vectores.  Por ejemplo, el coseno entre los vectores de palabras que tienen una similitud semántica importante será cercana a uno. De lo anterior se deduce que tales vectores ocupan posiciones cercanas dentro del campo semántico. Por el contrario, el coseno entre los vectores de palabras muy diferentes será cercano a cero, indicando una distancia considerable entre ellos, y por lo tanto una diferencia semántica importante entre ellas.  \\

\subsubsection*{7.Comparación semántica de textos}

En Günther, Dudschig $\&$ Kaup (2014) se propone una extensión del presente algoritmo, la cual se utiliza para la comparación de textos compuestos por dos o más palabras.  Lo anterior a través calcular los vectores propios promedios correspondientes a cada párrafo que se desea contrastar.  Es decir, sean $\overrightarrow{x}_{1}$,...,$\overrightarrow{x}_{p}$ los vectores propios de cada una de las palabras que componen el párrafo A, entonces su vector promedio está dado por $\overline{x}_{A}= \frac{ \sum_1^p  \overrightarrow{x}_{i}}{p}$. Por otra parte, el vector promedio de los valores propios del párrafo B se define como $\overline{y}_{B}= \frac{ \sum_1^n \overrightarrow{y}_{i}}{n}$ para $\overrightarrow{y}_{1}$,...,$\overrightarrow{y}_{n}$. De esta manera, dos párrafos tendrán un significado semántico parecido, en función de la cercanía entre los vectores  $\overrightarrow{x} _{A}$  y  $\overrightarrow{y} _{B}$. Así también, obtenemos el coseno del ángulo entre los vectores de interés.  Un coseno igual a 0 entre dos palabras nos indicará ortogonalidad, y por lo tanto, una diferencia semántica importante entre ambas. Por el contrario, entre más cercano a 1.0 sea el coseno entre vectores, nos indicará una similitud semántica mayor.\\

\subsubsection*{Paquetes para implementar el LSA}

El paquete semantics vectors propuesto por Widdows $ \&$ Cohen (2010) fue creado en el ambiente de JAVA y utiliza la técnica de proyecciones aleatorias para construir espacios semánticos. Este paquete tiene funciones básicas para calcular cosenos entre palabras individuales o párrafos.  El paquete DISSECT (DIstributional SEmantics Composition Toolkit) se está diseñado sobre el lenguaje de programación Python y permite crear espacios semánticos a partir de matrices documentos-palabras. Así también, puede ser utilizado para calcular similitudes semánticas entre palabras, listas de palabras o párrafos. $(ver: http://clic.cimec.unitn.it/composes/toolkit/)$

El campo semántico que se presenta en este trabajo fue realizado con el paquete LSA (Wild, 2015), el cual fue realizado dentro del ambiente de programación R. \\

\section*{Resultados }

En esta sección se presentan los resultados obtenidos a través de la implementación de la metodología explicada en la sección anterior.  Es importante considerar que la visualización correspondiente a la Figura 3 fue obtenida a partir de la técnica de K-Vecinos Cercanos (KNN, por sus siglas en inglés). Fue propuesta por Aha et. al. (1991), como una forma sencilla de predecir o clasificar nuevos datos, con base en observaciones pasadas o conocidas. Lo anterior en función a dos tipos de distancias: Euclidiana y de Mahalanobis, tal como se ilustra en las siguientes formulaciones:\\

\begin{equation}
D_{euclideana}\big(\overrightarrow{x},\overrightarrow{y}\big)=\sqrt{(\overrightarrow{x}-\overrightarrow{y})^{T}(\overrightarrow{x}-\overrightarrow{y})} 
\end{equation}\\

\begin{equation}
D_{mahalanobis}\big(\overrightarrow{x},\overrightarrow{y}\big)=  \sqrt{(\overrightarrow{x}-\overrightarrow{y})^{T} \sum\nolimits_{-1}^{}(\overrightarrow{x}-\overrightarrow{y})}
\end{equation}\\

Esta fuera del alcance de este trabajo profundizar en este tema y una explicación detallada sobre la técnica de KNN se puede consultar en Aha et al (1991); Weinberger y Saul (2009);  Muja y Lowe (2009). La siguiente parte de esta sección se enfoca en ilustrar un par de ejemplos numéricos.  
El primer ejemplo consiste en contrastar la similitud semántica entre las palabras “ciencia” y “matemáticas”, el cual es igual a 0.5988, así también las palabras “imaginación” y “poesía” registraron una similitud de 0.5528. Por otra parte, se reporta una similitud igual a 0.0063 entre las palabras “vaca” y “matemáticas”. Este valor es igual a 0.03810 para las palabras “toro” y “matemáticas”. Nótese que la combinación “toro-matemáticas” es seis veces mayor que “vaca-matemáticas”. Lo anterior se explica porque la palabra toro, si bien denota a un mamífero rumiante masculino del grupo de los bovinos, también hace referencia a la superficie generada por una circunferencia que gira alrededor de una recta exterior coplanaria. Con estos ejemplos se ilustra la forma de cuantificar similitudes y diferencias semánticas entre palabras, las cuales pueden aparentar no tener ningún tipo de relación a primera vista, pero existen de forma latente, como es el caso de las palabras “toro-matemáticas”. \\\\

\lstset{language=R, breaklines=true, basicstyle=\footnotesize}
\lstset{numbers=right, numberstyle=\tiny, stepnumber=1, numbersep=-2pt}
\begin{lstlisting}[frame=single]
# similitud semantica entre ciencia y matematicas 
> Cosine('ciencia','matematicas',tvectors=D)
[1] 0.5988701
# similitud semantica entre imaginacion y poesia 
> Cosine('imaginacion','poesia',tvectors=D)
[1] 0.5528838
# similitud semantica entre matematicas y toro 
> Cosine('matematicas','toro',tvectors=D)
[1] 0.03810167
# similitud semantica entre matematicas y vaca 
> Cosine('matematicas','vaca',tvectors=D)
[1] 0.00632693
\end{lstlisting}

El segundo caso se enfoca en comparar grupos de palabras. Partiendo de un vector específico es posible identificar a sus n vecinos más cercanos dentro del campo semántico. Es decir, estos vecinos estarán dados por aquellos n vectores que tienen la menor distancia con respecto al vector de referencia.  La interpretación es la siguiente: la similitud semántica entre el grupo de n vectores (o palabras) es inversamente proporcional a su distancia dentro del campo semántico. En la Figura 3 \label{Vecinos mas cercanos a "Matemáticas"} aparecen los 29 vecinos más cercanos a la palabra “matemáticas”. En este caso, las distancias están dadas por las primeras dos dimensiones del campo semántico, las cuales a su vez corresponden a los mayores valores propios. Nótese que “matemáticas” aparece en el origen del plano cartesiano y las palabras más próximas son “geometría”, “computacional”, “formulación” y “teoría”. \\\\

\begin{figure}[H]
\centering
\includegraphics[width=.60\textwidth]{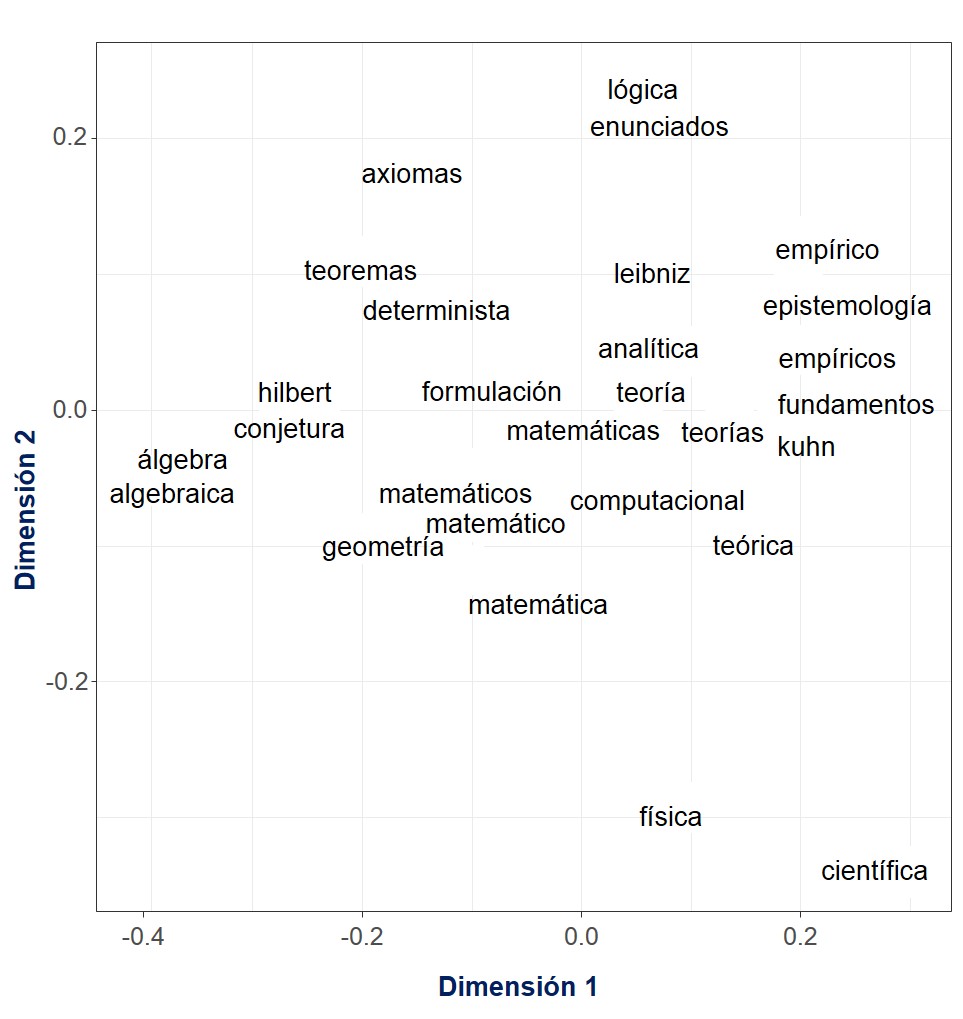}
\caption{Vecinos mas cercanos a "Matemáticas"}
\label{Vecinos mas cercanos a "Matemáticas"}
\end{figure}

En la última parte de los resultados se ilustran las comparaciones semánticas entre textos. Para tal propósito, hemos tomado tres párrafos con contextos aparentemente parecidos. El primero de ellos hace referencia al Artículo I de la Constitución Política de los Estados Unidos Mexicanos (EUM).  En segundo lugar, la Declaración Universal de los Derechos Humanos (DUDH) tomada del sitio de internet de las Organización de las Naciones Unidas (ONU, 2018) y, por último, la definición Derechos Humanos (DH.Wiki) según la Wikipedia (2018). 
A diferencia del ejemplo anterior, la comparación entre textos está dada por el promedio del grupo de palabras de los integran.  En primer lugar, el promedio de las distancias entre los vectores (palabras) que componen el primer texto (en este caso EUM) es obtenido, seguido del promedio del segundo texto (DUDH).  Posteriormente, se calcula el coseno entre los vectores resultantes de ambos grupos de palabras.  Este último representa el grado de similitud semántica entre los dos textos.  La similitud semántica entre el Artículo I de la Constitución Política de los Estados Unidos Mexicanos y la Declaración Universal de los Derechos Humanos (DUDH) es igual a 0.1523.  Lo anterior contrasta con el valor obtenido entre la definición de Derechos Humanos de la Wikipedia (DH.wiki), por una parte, y nuevamente el Artículo I (EUM) por la otra, el cual fue igual a 0.3315.  Además, comparación entre la DUDH y la definición de DH de la Wikipedia es igual a 0.3090.\\\\

\lstset{language=R, breaklines=true, basicstyle=\footnotesize}
\lstset{numbers=right, numberstyle=\tiny, stepnumber=1, numbersep=-2pt}
\begin{lstlisting}[frame=single]
# Comparacion de texto entre EUM y DUDH 
> Costring(EUM,DUDH,tvectors=D)
[1] 0.1523499
# Comparacion de texto entre EUM y DH.wiki
> Costring(EUM,DH.wiki,tvectors=D)
[1] 0.3363329
# Comparacion de texto entre DUDH y DH.wiki 
> Costring(DUDH,DH.wiki,tvectors=D)
[1] 0.3090089
\end{lstlisting}

Estos resultados tienen interpretaciones interesantes. En primer lugar, los textos EUM y DH.wiki presentan la mayor similitud (0.336). Por el contrario, la diferencia más importante se registró entre los textos EUM y DUDH (0.152). Si bien la similitud semántica entre ambos documentos oficiales, por una parte, y la definición genérica de la Wikipedia por la otra, es relativamente similar (0.336 y 0.309), llama la atención que este valor baje hasta 0.152 para la comparación entre ellos (EUM y DUDH). Hemos encontrado que el artículo I de la Constitución política de los EUM es semánticamente más parecido a la definición de DH de Wikipekia que la declaración de los derechos humanos de las Naciones Unidas.  De esta forma, podemos identificar diferencias semánticas entre dos más textos, los cuales principio están tocando el mismo tema. \\

\begin{figure}[H]
\centering
\includegraphics[width=.60\textwidth]{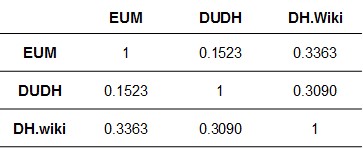}
\caption{Comparación semántica de textos}
\label{Comparación semántica de textos}
\end{figure}

 \subsubsection{Discusion}
En este trabajo se aplica un tipo de método estadístico multivariante denominado descomposición en valores singulares, con la finalidad de construir un campo semántico. De acuerdo con Dumais (2005), partiendo un dominio textual de gran envergadura, un campo semántico consiste en conceptualizar cada palabra de este dominio como un vector.  En este arreglo de alta dimensionalidad, aquellas palabras con significado similar ocuparan espacios cercanos dentro del espacio. Debido a que es necesario procesar bases de datos de gran tamaño para que se cumpla la hipótesis distribucional descrita en Hofmann (1999), la demostración numérica de tal hipótesis se realizó recientemente, en parte como consecuencia del incremento de la capacidad procesamiento que experimentaron las computadoras en los últimos años.  Si bien los campos semánticos han sido ampliamente estudiados para el idioma inglés, para el español el número de estudios es limitado. De esta forma, al proponer un campo semántico que describe las relaciones entre la mayoría de las palabras que componen el idioma español, buscamos hacer una contribución novedosa. \\

El campo semántico aquí propuesto puede ser utilizado eficientemente en comparar palabras o textos tomando como referencia el significado de cada unidad lingüística. De acuerdo con Furnas et. al (1987), la forma en que los seres humanos pueden utilizar y ordenar las palabras es un fenómeno complejo y este trabajo está limitado en tal dirección.  El trabajo propuesto por Bradford (2005), el cual propone versión revisada del análisis semántico basada en índices, hace una contribución para sopesar esta limitante.  Desde la perspectiva de la psicología Landauer. et. al (2007) menciona que una persona, desde su niñez hasta su vida adulta, se expone a cientos de millones de palabras. Estas serían las dimensiones de la matriz necesaria para poder construir un campo semántico, el cual pudiera describir realmente como los humanos utilizan el lenguaje. Es posible que veamos computadoras capaces de procesar estos volúmenes de información en un futuro cercano.

\section*{Referencias}
\begin{list}{•}{}

\item Aha, D. W., Kibler, D., $\&$ Albert, M. K. (1991). Instance-based learning algorithms. \textit{ Machine Learning }, 6(1), 37-66. \\
DOI:$\underline{10.1007/bf00153759}$

\item Attardi, G., Cozza, V., $\&$ Sartiano, D. (2015). Annotation and Extraction of Relations from Italian Medical Records. \textit{IIR.}

\item Barahona, I. (2018). Poverty in Mexico: Its relationship to social and cultural indicators. \textit{ Social Indicators Research }. 135(2), 599-627. \\
DOI: $\underline{1007/s11205-016-1510-3}$

\item Berry, M. W. (1992). Large-scale sparse singular value computations.\textit{ The International Journal of Supercomputing Applications}, 6(1), 13-49.

\item Bradford, R. (2005). Efficient discovery of new information in large text databases. In \textit{International Conference on Intelligence and Security Informatics }(pp. 374-380). Springer, Berlin, Heidelberg.

\item Cuadras, C. M. (2014). Nuevos Métodos De Análisis Multivariante. CMC Editions, 304.

\item Deerwester, S., Dumais, S. T., Furnas, G. W., Landauer, T. K., $ \&  $ Harshman, R. (1990). Indexing by latent semantic analysis. \textit{ Journal of the American society for information science}, 41(6), 391-407.

\item Dumais S. (2005). "Latent Semantic Analysis".\textit{ Annual Review of Information Science and Technology.} 38: 188–230.

\item Furnas, W., Landauer, K., Gomez, M., Dumais, S. (1987). "The vocabulary problem in human-system communication".\textit{ Communications of the ACM.} 30 (11): 964–971.

\item Günther, F., Dudschig, C., $ \& $ Kaup, B. (2015). LSAfun-An R package for computations based on Latent Semantic Analysis.\textit{ Behavior research methods}, 47(4), 930-944.

\item Hofmann T. (1999). "Probabilistic Latent Semantic Analysis".\textit{Uncertainty in Artificial Intelligence.}

\item Kintsch W $\&$ Bowles A. (2002) Metaphor Comprehension: What Makes a Metaphor Difficult to Understand?, Metaphor and Symbol, 17:4, 249-262,\\
DOI: $ \underline{10.1207/S15327868MS1704 1}$

\item Kintsch W .(1998) The Representation of Knowledge in Minds and Machines, International Journal of Psychology, 33:6, 411-420, DOI: $ \underline{10.1080/002075998400169}$

\item Kintsch, W. (2001). Predication. \textit{ Cognitive science}, 25(2), 173-202.

\item Lanczos, C. (1950). \textit{An iteration method for the solution of the eigenvalue problem of linear differential and integral operators.} 
Los Angeles, CA: United States Governm. Press Office.

\item Landauer T .(1999) Latent semantic analysis: A theory of the psychology of language and mind, \textit{ Discourse Processes}, 27:3, 303-310, DOI: $\underline{ 10.1080/01638539909545065}$

\item Landauer, T. K.,$ \&$ Dumais, S. T. (1997). A solution to Plato's problem: The latent semantic analysis theory of acquisition, induction, and representation of knowledge. \textit{ Psychological review}, 104(2), 211.

\item Landauer, T., et al., Handbook of Latent Semantic Analysis, Lawrence Erlbaum Associates, 2007.

\item Muja, M., $\&$ Lowe, D. G. (2009). Fast approximate nearest neighbors with automatic algorithm configuration. 
\textit{VISAPP (1)}, 2(331-340), 2.

\item Nakov, P., Popova, A., $ \&$ Mateev, P. (2001). Weight functions impact on LSA performance. EuroConference RANLP, 187-193.
ONU. (2018). Declaración Universal de los Derechos Humanos.
 Retrieved from \\ $\underline{http://www.un.org/es/sections/what-we-do/protect-human-rights/}$,\\ September $16^{th}$, 2018

\item Organización de las Naciones Unicas (ONU). (2018). \textit{Declaración Universal de los Derechos Humanos}. 
Accesado el 10 de octubre de 2018 de: \\$\underline {http://www.un.org/es/sections/what-we-do/protect-human-rights/}$\\

\item Rohde D,. Berry M,. Do, T,. O'Brien, G,. Krishna V, Varadhan, S,.  $ \&$ Evert S,. (2018). sparsesvd: Sparse Truncated Singular Value Decomposition (from 'SVDLIBC'). R package version 0.1-4. $ \underline{https://CRAN.R-project.org/package=sparsesvd}$

\item Weinberger, K. Q., $\&$ Saul, L. K. (2009). Distance metric learning for large margin nearest neighbor classification. 
\textit{Journal of Machine Learning Research}, 10(Feb), 207-244.

\item Widdows, D.,$ \&$ Cohen, T. (2010, September). The semantic vectors package: New algorithms and public tools for distributional semantics. In \textit{Semantic computing (icsc), 2010 ieee fourth international conference on }(pp. 9-15). IEEE.

\item Wikiextracto .(2018). Extracts and cleans text from a Wikipedia database dump. 
Retrieved from $ \underline{ https://github.com/attardi/wikiextractor}$

\item Wikipedia. (2018). Descargar Wikipedia Español. \\
Retrieved from $ \underline{ https://es.wikipedia.org/wiki/Wikipedia:Descargas} $, September 20th, 2017

\item Wikipedia. (2018). Definición de derechos humanos.\\
 Retrieved 2018, from $\underline{ https://es.wikipedia.org/wiki/Derechos_humanos.}$, October, 11th, 2018

\item Wikipedia. (2018). Free online encyclopedia.\\
 Retrieved from $\underline{ https://es.wikipedia.org/wiki/Wikipedia}$,  December 3rd, 2018

\item Wild F. (2015). lsa: Latent Semantic Analysis. R package version 0.73.1.\\
$ \underline{ https://CRAN.R-project.org/package=lsa}$

\end{list}

\end{document}